\documentclass{llncs}

\usepackage{graphicx}
\usepackage{amsmath,amssymb} 
\usepackage{color}

\usepackage{bm}
\usepackage{subfigure} 
\usepackage{epstopdf}
\usepackage{tabularx}
\usepackage{algorithmic}
\usepackage{algorithm}
\usepackage{hhline}
\usepackage{booktabs}
\usepackage{multirow}
\usepackage{arydshln}
\newcolumntype{C}{>{\centering\arraybackslash}X}
\usepackage[width=122mm,left=12mm,paperwidth=146mm,height=193mm,top=12mm,paperheight=217mm]{geometry}
\begin{document}
\pagestyle{headings}
\mainmatter

\title{Dense Light Field Reconstruction From Sparse Sampling Using Residual Network} 



\author{Mantang Guo, Hao Zhu, Guoqing Zhou, Qing Wang \\qwang@nwpu.edu.cn}
\institute{Northwestern Polytechnical University, Xi'an, China}

\maketitle

\begin{abstract}
A light field records numerous light rays from a real-world scene. However, capturing a dense light field by existing devices is a time-consuming process. Besides, reconstructing a large amount of light rays equivalent to multiple light fields using sparse sampling arises a severe challenge for existing methods. In this paper, we present a learning based method to reconstruct multiple novel light fields between two mutually independent light fields. We indicate that light rays distributed in different light fields have the same consistent constraints under a certain condition. The most significant constraint is a depth related correlation between angular and spatial dimensions. Our method avoids working out the error-sensitive constraint by employing a deep neural network. We solve residual values of pixels on epipolar plane image (EPI) to reconstruct novel light fields. Our method is able to reconstruct 2 to 4 novel light fields between two mutually independent input light fields. We also compare our results with those yielded by a number of alternatives elsewhere in the literature, which shows our reconstructed light fields have better structure similarity and occlusion relationship.


\end{abstract}

\section{Introduction}

A dense light field contains detailed multi-perspective information of a real-world scene. Utilizing these information, previous work has demonstrated many exciting applications with light fields, including changing the focus \cite{ng2006digital}, depth estimation \cite{wanner2012globally,wang2016depth,jeon2015accurate,zhu2016efficient}, and saliency detection \cite{li2014saliency}. However, it is difficult for existing devices to properly capture such a large quantity of information. In early light field capturing methods, light fields are recorded by multi-camera arrays or light field gantries \cite{wilburn2005high} which are bulky and expensive. In recent years, commercial light field cameras such as Lytro \cite{Lytro} and Raytrix \cite{RayTrix} are introduced to the general public. But they are still unable to efficiently sample a dense light field due to their trade-off between angular and spatial resolution.

Many methods have been proposed to synthesize novel views using a set of sparsely sampled views in light field \cite{wanner2014variational,Zhang2015lightfield,kalantari2016learning,wu2017light}. But, these methods only increase the view density in a single light field. Kalantari et al. \cite{kalantari2016learning} proposed a learning-based method to synthesize novel view at arbitrary position in a light field by using views in the four corners of light field. Recently, Wu et al. \cite{wu2017light} proposed a leaning-based method to synthesize novel views by increasing the resolution of EPI. These methods outperform other state-of-the-art methods \cite{wanner2014variational,Zhang2015lightfield} on view synthesis. However, all these methods are only able to synthesize novel views in a single light field. Besides, in these methods, the baseline between sampled views has to be close enough. They cannot properly reconstruct a large quantity of novel light rays with wide baseline. 

In this paper, we explore dense light field reconstruction from sparse sampling. We propose a novel learning-based method to synthesize a great number of novel light rays between two distant input light fields, whose view planes are coplanar. Using the disparity consistency between light fields, we first model the relationship between EPIs of dense and sparse light field. Then, we extend the error-sensitive disparity consistency between EPIs in sparse light field by employing ResNet\cite{he2016deep}. Finally, we reconstruct a large quantity of light rays between input light fields. The proposed method is capable of rendering a dense light field by using multiple input light fields which are captured by commercial light field camera. In addition, the proposed method requires neither depth estimation nor other priors. Experimental results on real-world scenes demonstrate the performance of our proposed method. The proposed method is at most capable of reconstructing four novel light fields between two input light fields. Besides, in terms of the quality of synthesized novel view images, our method outperforms state-of-the-art methods on both quantitative and qualitative results.

Our main contributions are:

1) We present a learning-based method for reconstructing a dense light field by using a sparse set of light fields sampled by commercial light field camera.

2) Our method is able to reconstruct a large quantity of light rays and occlusions between two distant input light fields.

3) We introduce a high-angular-resolution light field dataset whose angular resolution is the highest among light field benchmark datasets so far.

\section{Related Work}
Dense sampled light field is in need for many computer vision applications. However, it costs much time and space to acquire and store massive light rays by existing devices and algorithms. Many research groups have focused on increasing a camera-captured light field's resolution by using a set of samples \cite{wanner2014variational,Zhang2015lightfield,kalantari2016learning,wu2017light,levin2010linear,marwah2013compressive,shi2014light,schedl2015directional,yoon2015learning,zhang2017plenopatch,srinivasan2017learning}. Here, we survey some state-of-the-art methods.
\subsection{View-based vs. EPI-based Angular Interpolation}
Wanner and Goldluecke \cite{wanner2014variational} used the estimated depth map to warp input view image to novel view. However, the quality of synthesized view is easily affected by the accuracy of depth map. Levin et al. \cite{levin2010linear} used a new prior to render a 4D light field from a 3D focal stack. Shi et al. \cite{shi2014light} took advantage of the sparsity of light field in continuous Fourier domain to reconstruct a full light field. The method sampled multiple 1D viewpoint trajectories with special patterns to reconstruct a full 4D light field. Zhang et al. \cite{Zhang2015lightfield} introduced a phase-based method to reconstruct a full light field from micro-baseline image pair. Schedl et al. \cite{schedl2015directional} reconstructed a full light field by searching for best-matching multidimensional patches within the dataset. However, in these methods, due to the limitation of specific sampling pattern and algorithm complexity, they are unable to properly generate a dense light field. The angular resolution of synthesized light field is at most 20*20 with commercial light field cameras. Marwah et al. \cite{marwah2013compressive} proposed a method to reconstruct light field from a coded 2D projection. But they need a special designed equipment to capture compressive light field.

Recently, learning-based methods are explored in light field super-resolution. Kalantari et al. \cite{kalantari2016learning} introduced a learning-based method which used four corner view images to synthesize an arbitrary view image in a single light field. They used two sequential networks to estimate depth and color values of pixels in novel view image. Srinivasan et al. \cite{srinivasan2017learning} proposed a learning-based method to synthesize a full 4D light field by using a single view image. However, these methods heavily rely on the accuracy of depth map. Yoon et al. \cite{yoon2015learning} trained several CNNs to increase spatial and angular resolution simultaneously. However, the method could only synthesize one novel view between two or four input views. Wang et al. \cite{wang2017light} proposed a learning-based hybrid imaging system to reconstruct light field video. Although the work did not directly aim at novel view synthesis, it in fact had synthesized novel frames containing different views of a light field. In their proposed system, DSLR provided the prior information that is equivalent to the central view of each synthesized light field frame. Instead of using extra prior to guide light field reconstruction, our proposed method only use light fields which captured by commercial light field camera as input.

Apparently, EPI has a strong characteristic of linearity. Many methods explored light field processing based on EPI. However, there are fewer work focusing on angular interpolation of light field. Wu et al. \cite{wu2017light} trained a residual-based network to increase angular resolution of EPI. They employ a 1D blur kernel to remove high spatial frequency in the sparsely sampled EPI before feeding to the CNN. Then, they carry out a non-blind de-blur to restore high spatial frequency details which are removed by the blur kernel. However, due to the limitation of blur kernel's size and interpolation algorithm in the preprocessing, the method also fails when the baseline of input views is wide.

\subsection{View Synthesis vs. Light Field Reconstruction}

Many methods focus on light field view synthesis, such as \cite{wanner2014variational,Zhang2015lightfield,kalantari2016learning,wu2017light,levin2010linear,marwah2013compressive,shi2014light,schedl2015directional,srinivasan2017learning}  . The most important insight is that all these methods synthesize novel views in internal of a light field. Due to the narrow baseline among views of existing commercial light field camera, all the input view images in these methods have high overlapping ratio between each other. These methods use redundant information between input view images. Therefore, they are only able to synthesize novel views within input views. On the contrast, we propose a novel method that can reconstruct multiple light fields between two input light fields instead of reconstructing views inside a single light field. Therefore, our proposed method is capable of reconstructing a dense light field by using a sparse set of input light fields. Besides, the method is able to reconstruct a mass of occlusions in reconstructed light fields without requiring depth estimation.

\begin{figure*}[hb]
\begin{center}
\centering
\includegraphics[width=\linewidth]{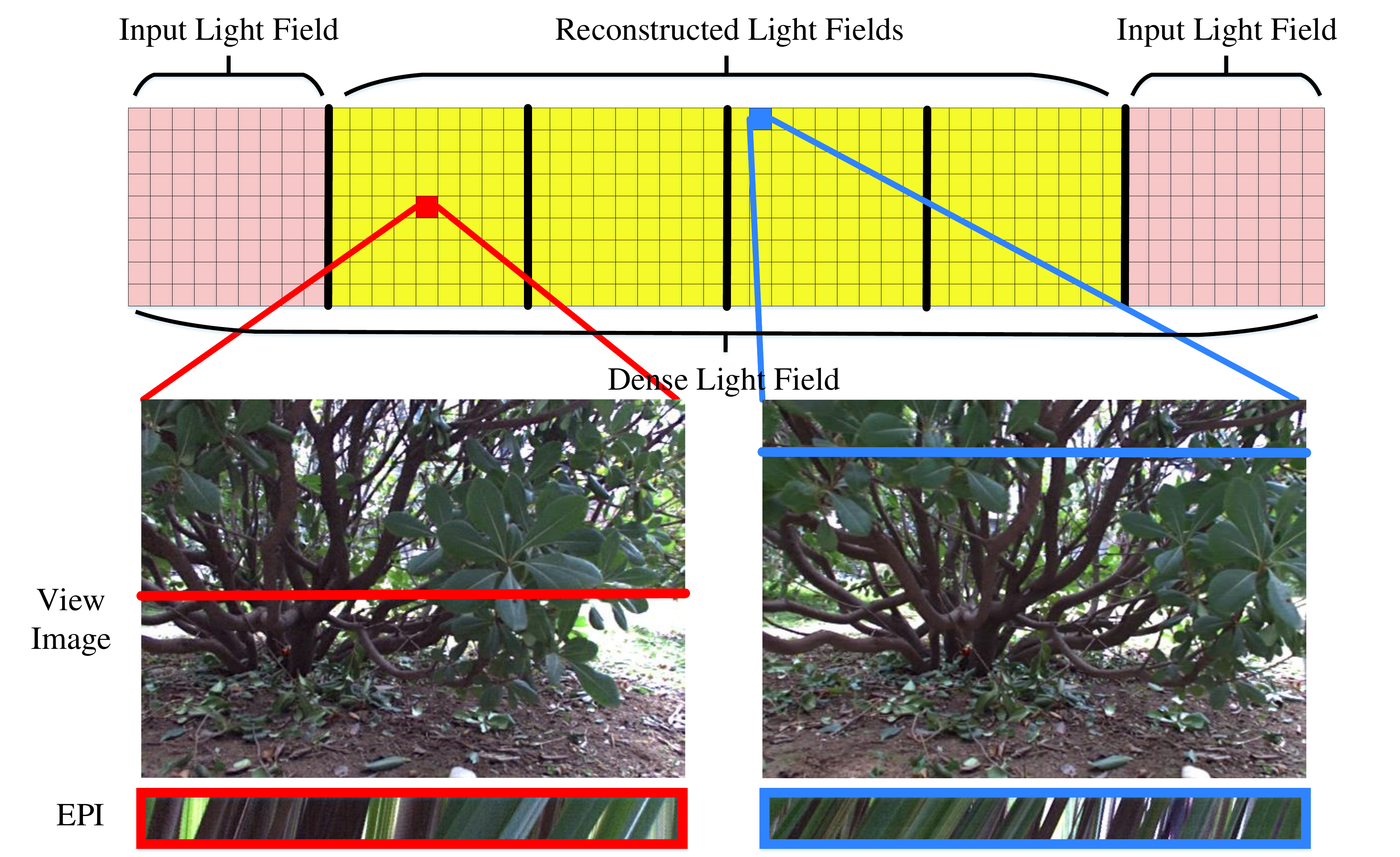}
\end{center}
\caption{An illustration of dense light field reconstruction (reconstruct 4 novel light fields between input light fields) from sparse sampling. The band on the top refers to the dense light field. In dense light field, input light fields are denoted as pink, while reconstructed light fields are denoted as yellow. Our method employs a ResNet to reconstruct multiple light fields between input light fields. The view images are from two of reconstructed light fields. The EPIs are from two horizontal lines (red and blue) in dense light field.}
\label{fig:Frontresult}
\end{figure*}

We model novel light field reconstruction based on 2D EPI. Different from other light field view synthesis methods, our proposed method directly reconstructs multiple novel light fields, as shown in Fig.~\ref{fig:Frontresult}. Besides, view planes of input light fields are not overlapping. There are hundreds of missing views between two input light fields. Therefore, the difficulty lies in that it needs to reconstruct a large amount of light rays and occlusions.

\section{Problem Formulation}
\begin{figure*}[tb]
\begin{center}
\centering
\includegraphics[width=\linewidth]{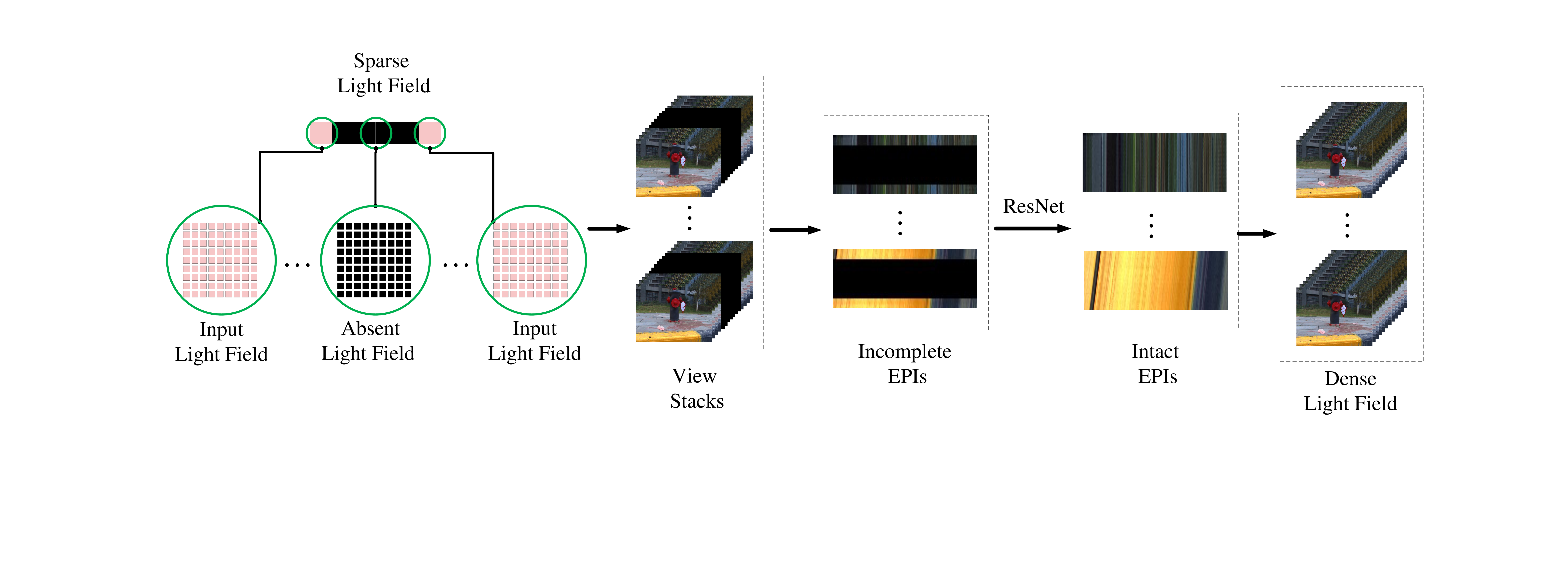}
\end{center}
\caption{The pipeline of the proposed method. Two input light fields' view planes are coplanar. The pixels' values in those absent light fields are initially set to zero.  By employing ResNet to predict the residual values between intact EPI and incomplete EPI, all those zeroth pixels can be predicted. }
\label{fig:Pipeline}
\end{figure*}

In the two-parallel-plane parameterization model, light field is formulated as a 4D function $L(u,v,s,t)$ \cite{levoy1996light}, where the pair $(u,v)$ represents the intersection of light ray and view plane, and $(s,t)$ represents the intersection of light ray and image plane. In the paper, we assume that input light fields' view planes are coplanar (see Fig.~\ref{fig:Pipeline}). Besides, their view planes are non-overlapping. Our task is to interpolate a great deal of novel light rays between these light fields.

All views in a light field are assumed as perspective cameras with identical intrinsic parameters. The transformation between views within the same light field is merely a translation without rotation. Therefore, a 3D point $\mathbf{X}=(X,Y,Z)^\top $ in real-world scene is mapped to the pixel $(s,t)$ in the image plane of light field as follows.

\begin{equation}\small
\lambda \left[ \begin{array}{c}
s \\ t \\ d^{(s,t)} \\ 1
\end{array} \right]=
\left[ \begin{array}{c@{\hspace{10pt}}c@{\hspace{10pt}}c@{\hspace{10pt}}c}
f & 0 & 0 & -fu \\
0 & f & 0 & -fv \\
0 & 0 & 0 & B \\
0 & 0 & 1 & 0 \\
\end{array} \right]
\left[ \begin{array}{c}
X \\ Y \\ Z \\1
\end{array} \right]
\label{eq:eq1}
\end{equation}
where $B$ denotes a constant, $d^{(s,t)}$ refers to the disparity of $(s,t)$ , $f$ is the interval between view plane and image plane. Then, the constraints of a light ray in light field can be described as

\begin{equation}\small
\left\lbrace\begin{aligned}
u &= X - \frac{B}{f}\frac{s}{d^{(s,t)}}\\
v &= Y - \frac{B}{f}\frac{t}{d^{(s,t)}}\\
\end{aligned}\right.
\label{eq:eq2}
\end{equation}

In fact, Eq.~\ref{eq:eq2} is the mathematical description of EPI which is a 2D slice cut from 4D light field. Its simple linear structure makes it easy to analyze in light field. As a specific representation of light field, EPI not only includes angular information but also contains spatial information. EPI mainly has two significant properties. One property is that a scene point is represented by a straight line whose slope is a constant value \cite{bolles1987epipolar}. Another property is that pixels on a straight line refer to different light rays emitting from the same point. In fact, the slope of a line in EPI reflects the disparity of a point observed in different views. We define this linear constraint between $u$ and $s$ in EPI as disparity consistency, as formulated in Eq.~\ref{eq:eq2}. Based on disparity consistency, a EPI can be further formulated as
\begin{equation}
epi(u,s)=epi(X - \frac{B}{f}\frac{s}{d^{(s,t)}}, s)
\label{eq:eq3}
\end{equation}

For any two light fields in our assumption, the transformation between them is merely a translation without rotation

\begin{equation}\small
\mathbf{X}'= [\mathbf{I}|\mathbf{t}]\mathbf{X}
\label{eq:eq4}
\end{equation}
where $\mathbf{I}$ is an identity matrix, $\mathbf{t}=(t_x,t_y,0)^\top$, $\mathbf{X}$ and $\mathbf{X}'$ are two scene points. Besides, the disparity of an identical scene point stays the same in multiple light fields. The transformation between EPIs of any two light fields in our assumption is formulated as
\begin{equation}
\begin{aligned}
epi'(u',s')&=epi'(X'- \frac{B}{f}\frac{s'}{d^{(s',t')}}, s')\\
&=epi(X+t_x-\frac{B}{f}\frac{s'}{d^{(s',t')}}, s')\\
&=epi(X-\frac{B}{f}\frac{s}{d^{(s,t)}}, s+\frac{f}{B}d^{(s,t)}t_x)
\end{aligned}
\label{eq:eq5}
\end{equation}
where $d^{(s',t')}$ is equal to $d^{(s,t)}$. Therefore, under the condition that two light fields' view planes are coplanar, their EPIs can represent each other through disparity consistency. 

 In our model, there are two kinds of light fields (see Fig.~\ref{fig:Pipeline}). One is the sparse light field which is made up by input light fields. The other one is the dense light field which is reconstructed based a sparse light field. In terms of the universe of light rays, the sparse light field is actually a subset of the dense light field. Besides, both of them meet the condition that their view planes are coplanar under our assumption. Therefore, their EPI can also represent each other through disparity consistency.

\begin{equation}\small
{E_{sparse}}(u',s') = {E_{dense}}(X - \frac{B}{f}\frac{s}{{{d^{(s,t)}}}},s + \frac{f}{B}{d^{(s,t)}}{t_x})
\label{eq:eq6}
\end{equation}
where $E_{dense}$ is the dense light field's EPI, $E_{sparse}$ is the EPI of sparse light field. Thus, we are able to reconstruct a dense light field by extending the disparity consistency in sparse light field.

\section{Reconstruction based on Residual Network}

For gaining disparity consistency, disparity estimation is an error-sensitive solution with existing algorithms. In our proposed method, we extend the disparity consistency among light fields by employing a neural network. 

Compared with dense light field, there are many light rays being absent in sparse light field which is composed by input light fields (see Fig.~\ref{fig:Pipeline}). Many entire rows of pixels are needed to be reconstructed in its EPI. These missing rows form a blank band in EPI. We initially set these pixels' values to zero in the blank band, as shown in Fig.~\ref{fig:Pipeline}. Our task is to find an operation that can predict the pixels' values in blank band.

\begin{figure*}[htb]
\begin{center}
\centering
\includegraphics[width=\linewidth]{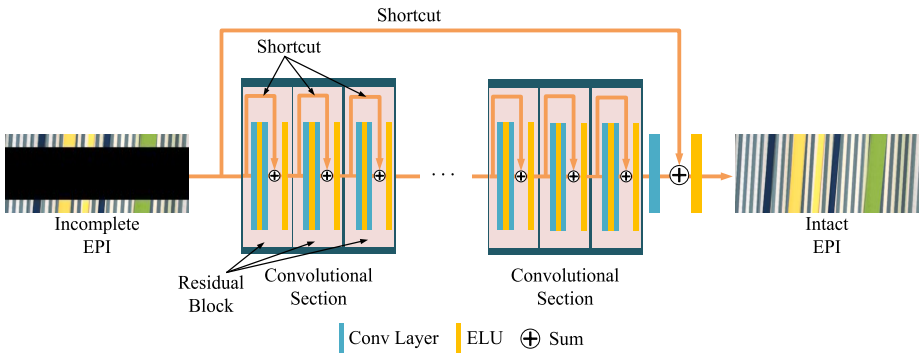}
\end{center}
\caption{The network contains 5 convolutional sections. Each section has 3 residual blocks which are defined in \cite{he2016deep}. From the first section to the fifth one, the number of filters and filter sizes are configured as (32, 9), (64, 7), (128, 5), (256, 5), (512, 5) respectively. The input and output of the network is incomplete RGB EPI and intact EPI respectively. We use shortcut operation to maintain high frequency details in EPI. Each convolutional layer is followed by an exponential linear unit (ELU). }
\label{fig:Resnet}
\end{figure*}

\subsection{Network Architecture}
In fact, pixels' values of input light fields remain unchanged in the sparse light field's EPI during the reconstructing procedure. We only need to predict pixels' values in blank band. Therefore, we regard pixels' values in blank band as the residual values between EPI of dense light field and EPI of sparse light field:
\begin{equation}\small
Res=E_{dense}-E_{sparse}
\label{eq:eq7}
\end{equation}
where $Res$ refers to the residual between $E_{dense}$ and $E_{sparse}$.  Thus, we employ ResNet\cite{he2016deep} to predict the residual. Besides, due to particular residual blocks and shortcuts in the network, the network only needs to consider the residual between input and output and preserves high frequency details of EPI. We reformulate the reconstruction of dense light field's EPI $E_{dense}$ as follows:
\begin{equation}\small
\mathop {\min }\limits_{res,\theta } \left\| {{{E_{sparse} + res(E_{sparse}}},\theta ) - E_{dense}} \right\|
\label{eq:eq8}
\end{equation}
where $res$ refers to the operation of residual network that solves residuals between input and output. $\theta$ refers to parameters of convolution layers in the network. Therefore, the residual between $E_{sparse}$ and $E_{dense}$ can be solved by minimizing the difference between output and ground truth iteratively, which refers to ${{{E_{sparse}}}+{{res(E_{sparse}}},\theta )}$ and $E_{dense}$ respectively in Eq.~\ref{eq:eq8}. 

The structure of supervised network is shown in Fig.~\ref{fig:Resnet}. The network contains 32 convolutional layers.The input and output are single RGB image of incomplete EPI and intact EPI. The main part of the network contains 5 convolutional sections, and each section has 3 residual blocks mentioned by He et al. \cite{he2016deep}. The layers in the same section have the same number of filters and filter size. In order to preserve high frequency details in EPI, we cancel the pooling operation throughout the network and maintain the input and output at the same size in each layer. 

\subsection{Training Details}
We have modelled light field reconstruction between input light fields as a learning-based end-to-end regression.  In order to minimize the error between the output of the network and ground truth, we use the mean squared error (MSE) as the loss function of our network,
\begin{equation}\small
L = \frac{1}{N}\sum\limits_{i = 1}^N {{{\left\| {\left. {{{\rm{E}}_{sparse}}^{(i)}{\rm{ + res(}}{{\rm{E}}_{sparse}}^{(i)},\theta ) - {{\rm{E}}_{dense}}^{(i)}} \right\|} \right.}^2}} 
\label{eq:eq9}
\end{equation}
where $N$ is the number of input EPIs. Since the training is a supervised process, we use EPI cut from the dense light fields in our dataset (see Section.~5) as ground truth to guide the training.

In the training process, in order to converge our training model efficiently and improve the accuracy, we initialize parameters of network's filter by using Xavier method~\cite{glorot2010understanding} and use the ADAM algorithm~\cite{kingma2014adam} to optimize the parameters. Besides, to prevent the model from overfitting, we augment training data by randomly adjusting the brightness of EPIs and adding Gaussian noise to EPIs. Furthermore, we train the network with 5 epochs and each epoch contains 2256 iterations. The learning rate is set 1e-4 initially. Then, it is decreased by a factor of 0.96 every 2256 iterations so as to make the model converge more quickly. There are 30 EPIs in each batch in the training process. The training of the network takes about 23 hours on 6 GPUs GTX 1080ti with the Tensorflow.

\section{Results}
In this section, we first explain the capturing process of our high-angular-resolution light field dataset. Then, we evaluate our proposed method on the light field dataset by using a sparse sampling pattern. In addition, we test our method's capacity of reconstructing dense light field with different sampling patterns.

\subsubsection{Dataset.}
The angular resolution of existing light field datasets is too low to verify our proposed method. Besides, our training is a supervised process. In order to provide ground truth dense light field during training process, we create a dense light field dataset. The dataset composes of 13 indoor scenes and 13 outdoor scenes. It contains plenty of real-world static scenes, such as bicycles, toys and plants, which have abundant colors and complicated occlusions. Each scene in the dataset contains 100 light fields captured by Lytro ILLUM. There are 2600 light fields in total.
\begin{figure*}[htb]
\begin{center}
\centering
\includegraphics[width=0.9\linewidth]{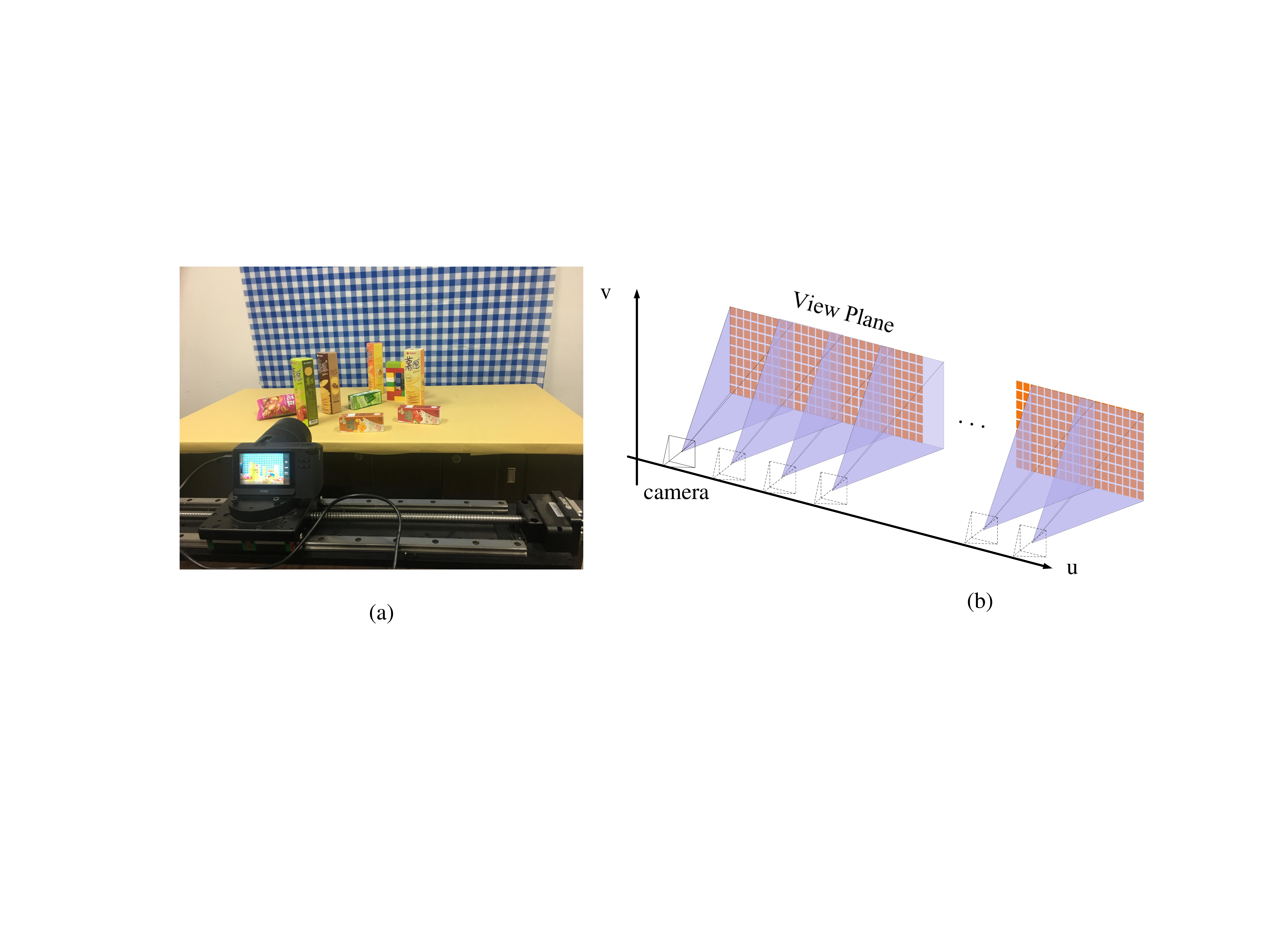}
\end{center}
\caption{(a) is the proposed capturing system of our method. The Lytro ILLUM is mounted on a translation stage in order to ensure all the captured light fields' view planes are coplanar; (b) models the capturing process. The camera move on a straight line with a proper step size while capturing light fields. The view planes of each pair of adjacent captured light fields has overlaps.}
\label{fig:devices}
\end{figure*}

For each scene, in order to make all the captured light fields' view planes coplanar, we mount a Lytro ILLUM on a translation stage and move the camera along a line in the capturing process. Our capturing system is shown in Fig.~\ref{fig:devices}(a). Furthermore, for the sake of gaining a dense light field from each scene, we set a proper step size for the translation stage during moving the camera. It ensures that there is overlap between each pair of adjacent light fields' view planes, as shown in Fig.~\ref{fig:devices}(b). In our experiment, there are 5 views that are overlapped between each pair of adjacent light fields. The camera focuses at infinity during the capturing. All the light fields are decoded by Lytro Power Tools~\cite{Lytro}. For each light field, central $9\times9$ views are extracted from $14\times14$ views provided by raw data to maintain the imaging quality. Then, we fuse the overlapping views between each pair of adjacent light fields to merge all the light fields together. After merging, each scene is recorded by a 405 high-angular-resolution light field. From another perspective, the high-angular-resolution light field is composed by 45 $9\times9$ low-angular-resolution light fields whose view planes connect with each other but have no overlapping views.

\begin{figure*}[tb]
\begin{center}
\centering
\includegraphics[width=\linewidth]{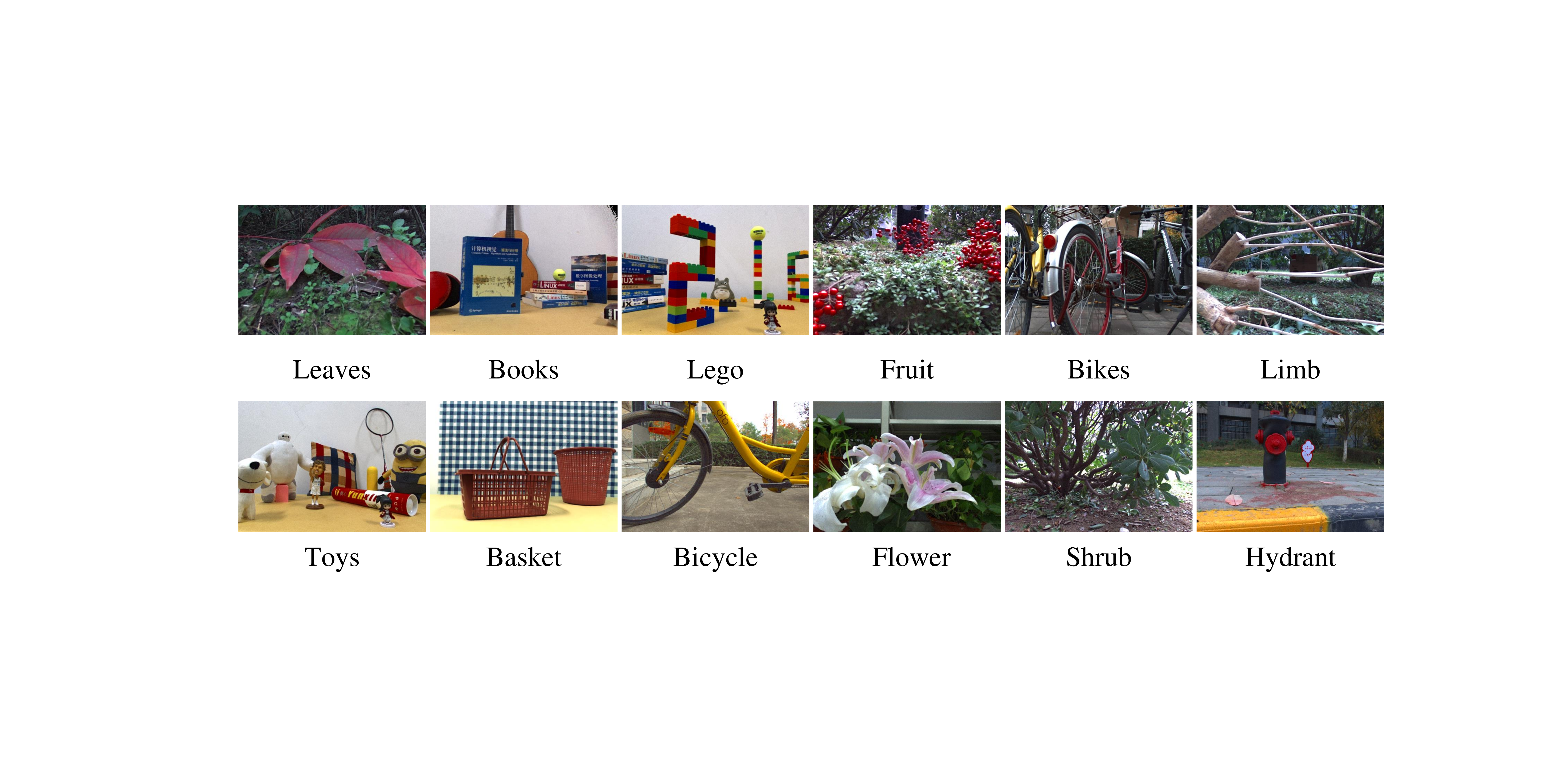}
\end{center}
\caption{The scenes in our light field dataset. The first row is a part of training data. The second row is testing data used to evaluate the proposed method.}
\label{fig:scenes}
\end{figure*}

\subsubsection{Real-world Scenes Evaluation }
\begin{figure*}[tb]
\begin{center}
\centering
\includegraphics[width=\linewidth]{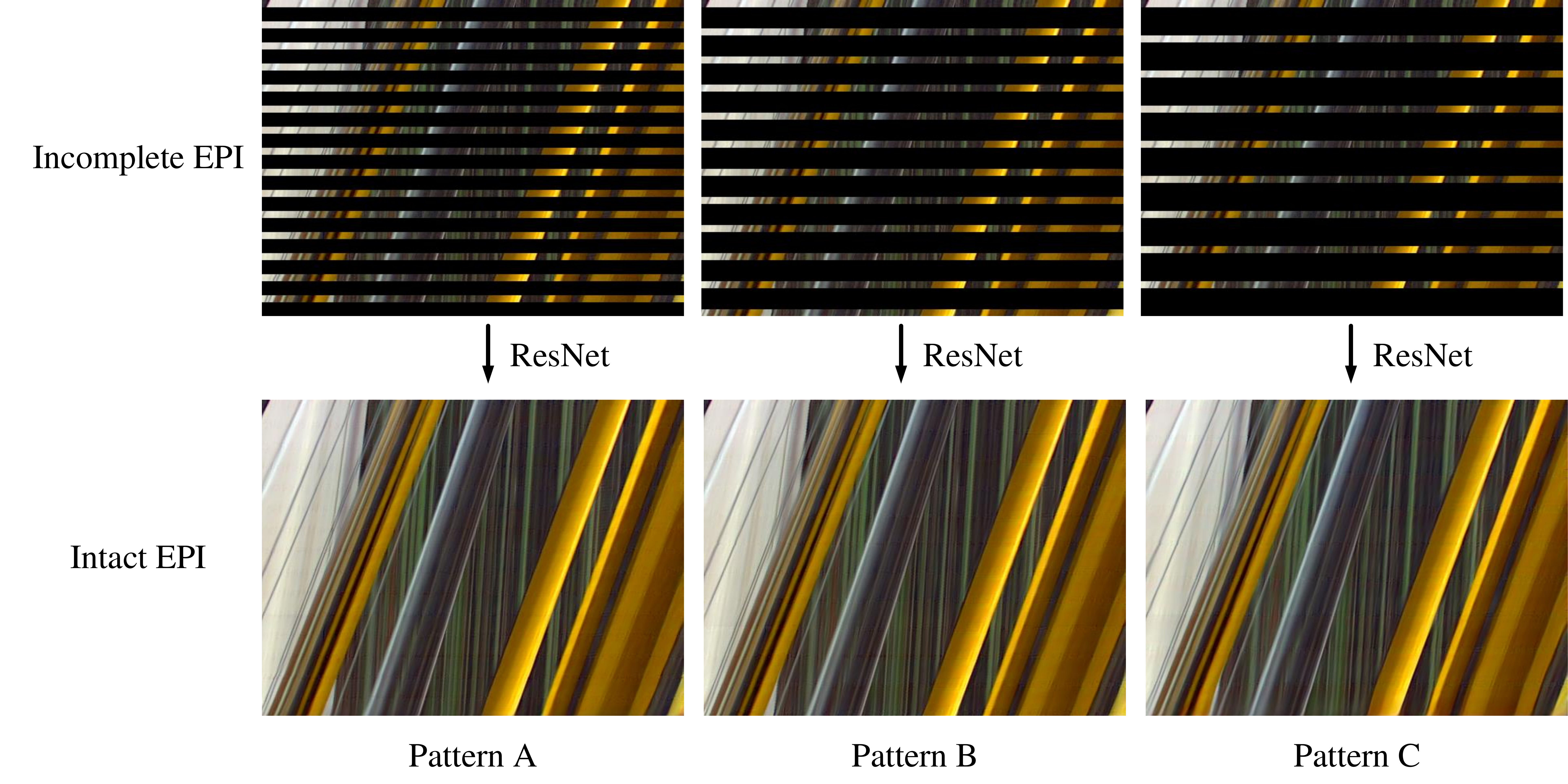}
\end{center}
\caption{The EPIs of sparse light field under different sampling patterns. The upper incomplete EPI with blank bands is cut from sparse light field. There is a certain number of light fields need to be reconstructed in blank band between each pair of input light fields. With incomplete EPI as input, our trained network outputs the lower intact EPI with all the blank bands filled in incomplete EPI. From pattern A to pattern C, the number of reconstructed light fields in each blank bands is respectively 2, 3, 4.}
\label{fig:samplepattern}
\end{figure*}

We design three sparse sampling patterns to evaluate the proposed method on our real-world dataset. With different sampling pattern, the number of light fields which need to be reconstructed between each pair of input light field is different. First, we sparsely sample multiple light fields in each scenes's dense light field to make up a sparse light field. Then, we use the EPI of sparse light field as our network's input to generate the intact EPI and reconstruct a dense light field. The sampling patterns are  shown in Fig.~\ref{fig:samplepattern}.  We choose 20 scenes as the training data which contains 67680 EPIs. The other 6 scenes (see Fig.~\ref{fig:scenes}) are used to test our training model and other methods. For our method, we reconstruct 2 novel $9\times9$ light fields between each pair of input light fields to verify the proposed method. The methods of Kalantari et al. \cite{kalantari2016learning} and Wu et al. \cite{wu2017light} perform better than other state-of-the-art methods. Thus, we use them to evaluate the quality of views in reconstructed light fields. When we evaluate these two methods on our real-world dataset, we carefully fine-tune all parameters so as to gain the best experimental performance among their results. Furthermore, we set the same up-sampling factor in their code.

The average PSNR and SSIM values are calculated on each testing scene's reconstructed view images, listed in Table.~\ref{tab:tab1}.
In the method of Kalantari et al. \cite{kalantari2016learning}, the quality of synthesized view is heavily dependent on the accuracy of depth map. It tends to fail in the Basket and Shrub data. Since these scenes are challenging cases for depth estimation. The method proposed by Wu et al. \cite{wu2017light} uses the ``blur-deblur" to increase the resolution of EPI instead of estimating depth. It achieves better performance than that of Kalantari et al. \cite{kalantari2016learning} on these. However, this method has to increase light field's angular resolution sequentially. The result of lower resolution-level's reconstruction is used as the input of higher resolution-level's reconstruction so that the constructing error is accumulated along with angular resolution's increasement. Our proposed method does not require error-sensitive depth estimation to reconstruct light field. Besides, all the light rays in the reconstructed light fields are synthesized at a time. Therefore, in terms of quantitative estimation, the results indicate that our proposed method is significant better than other methods on the quality of synthesized views.

\begin{table}[!tb]
\begin{center}
\centering
\caption{PSNR and SSIM results of reconstructed light fields on real-world scenes with pattern A. The values are averaged over all the views in reconstructed light fields.}
\label{tab:tab1}
\begin{tabularx}{.95\textwidth}{l|c|CCCCCC}
\hline
       & 	&Toys           & Basket         & Bicycle        & Flower         & Shrub                     & Hydrant        \\
\hline
            & Kalantari et al. \cite{kalantari2016learning}& 34.42          & 30.26          & 30.64          & 31.50          & 28.55          & 31.82          \\
PSNR        & Wu et al. \cite{wu2017light}       & 35.21          & 32.87          & 35.74          & 32.82          & 30.73          & 38.67          \\
            & Ours      & \textbf{40.78} & \textbf{40.46} & \textbf{39.25} & \textbf{38.55} & \textbf{34.81} & \textbf{40.75} \\
\hline
           	&Kalantari et al. \cite{kalantari2016learning}& 0.897          & 0.922          & 0.862          & 0.877          & 0.878          & 0.850 \\
SSIM       	&Wu et al. \cite{wu2017light}       & 0.919          & 0.958          & 0.943          & 0.904          & 0.910          & 0.947 \\
           	&Ours      & \textbf{0.942} & \textbf{0.987} & \textbf{0.949} & \textbf{0.943} & \textbf{0.932} & \textbf{0.958} \\
\hline
\end{tabularx}
\end{center}
\end{table}

\begin{figure*}[!tb]
\begin{center}
\centering
\includegraphics[width=\linewidth]{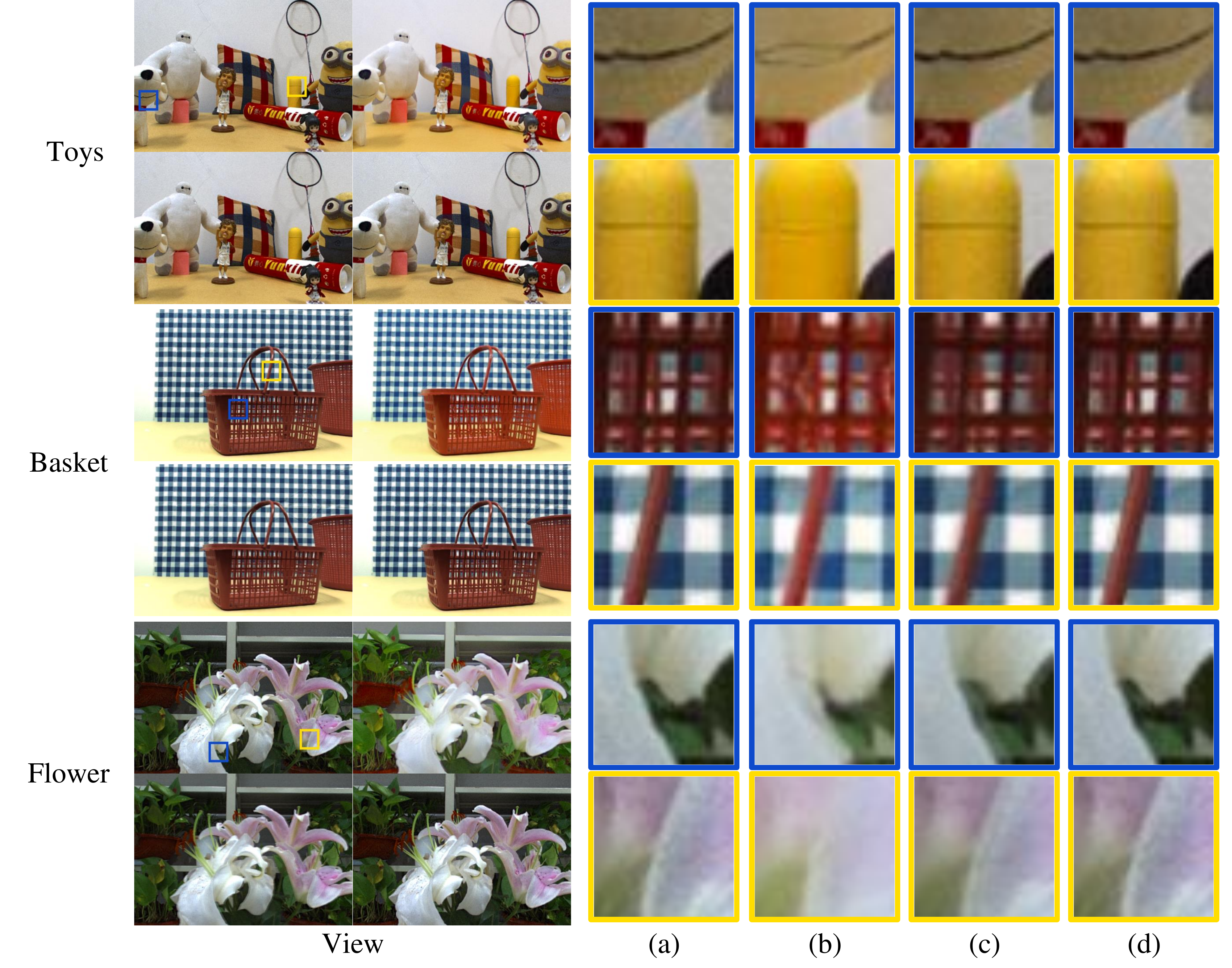}
\end{center}
\caption{The result of view images in reconstructed light field of 3 real-world scenes (reconstruct 2 novel light fields between each pair of input light fields). The first column shows view images. Upper left: ground truth. Upper right: Kalantari et al. \cite{kalantari2016learning}. Lower left: Wu et al. \cite{wu2017light}. Lower right: Ours.  From (a) to (d), the detailed results in the blue and yellow boxes are ground truth, Kalantari et al. \cite{kalantari2016learning}, Wu et al. \cite{wu2017light} and ours respectively.}
\label{fig:views}
\end{figure*}

\begin{figure*}[!t]
\begin{center}
\centering
\includegraphics[width=\linewidth]{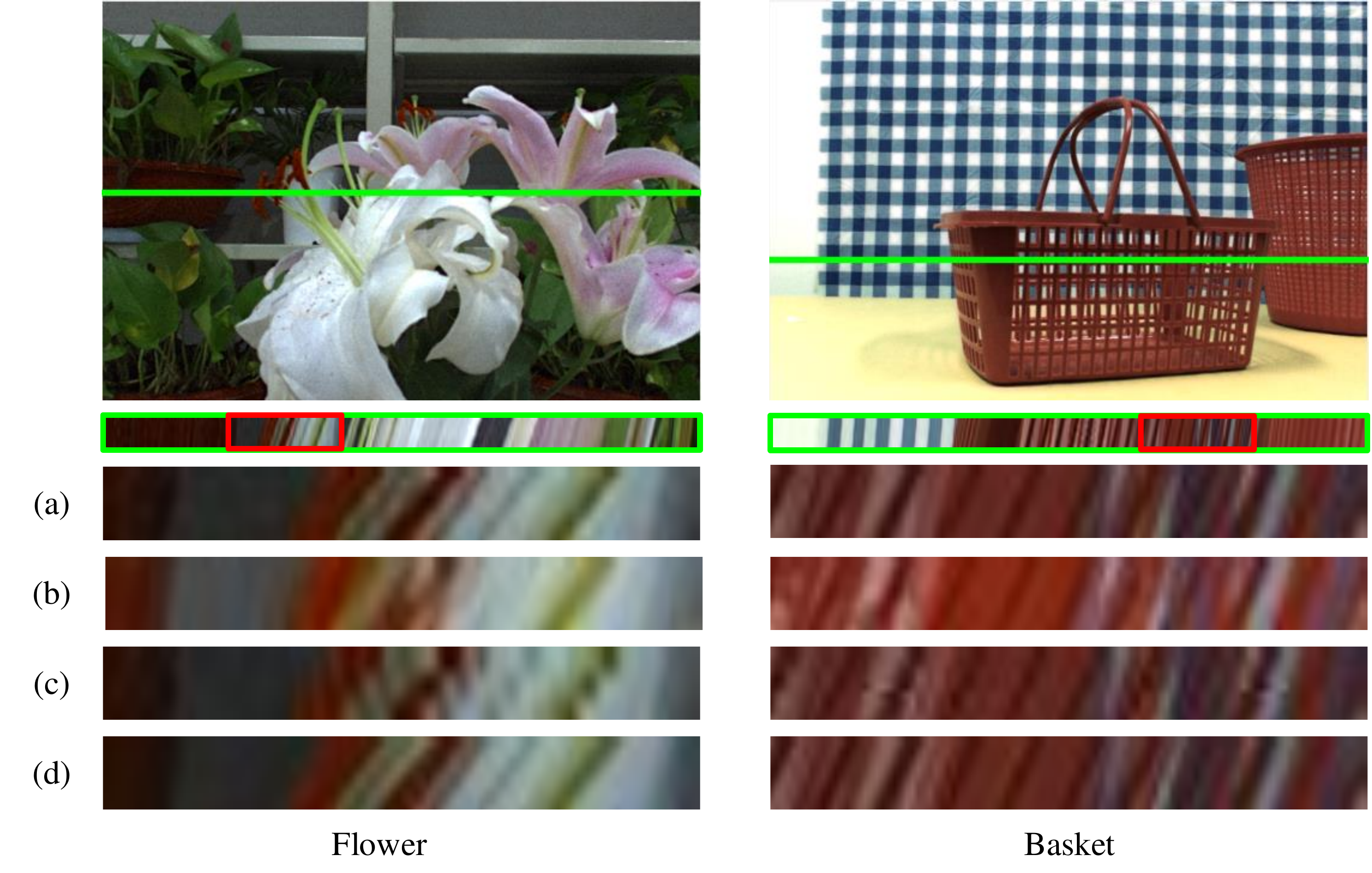}
\end{center}
\caption{The result of EPI of 2 real-world scenes. The EPIs enclosed by green box are the ground truth EPIs which are extracted from the green line in the view images. We up-sample each method's EPI result below for a better view. From (a) to (d), the detailed results are ground truth,  Kalantari et al. \cite{kalantari2016learning}, Wu et al. \cite{wu2017light} and ours respectively.}
\label{fig:EPIs}
\end{figure*}

\begin{figure*}[!t]
\begin{center}
		\centering
		\includegraphics[width=\linewidth]{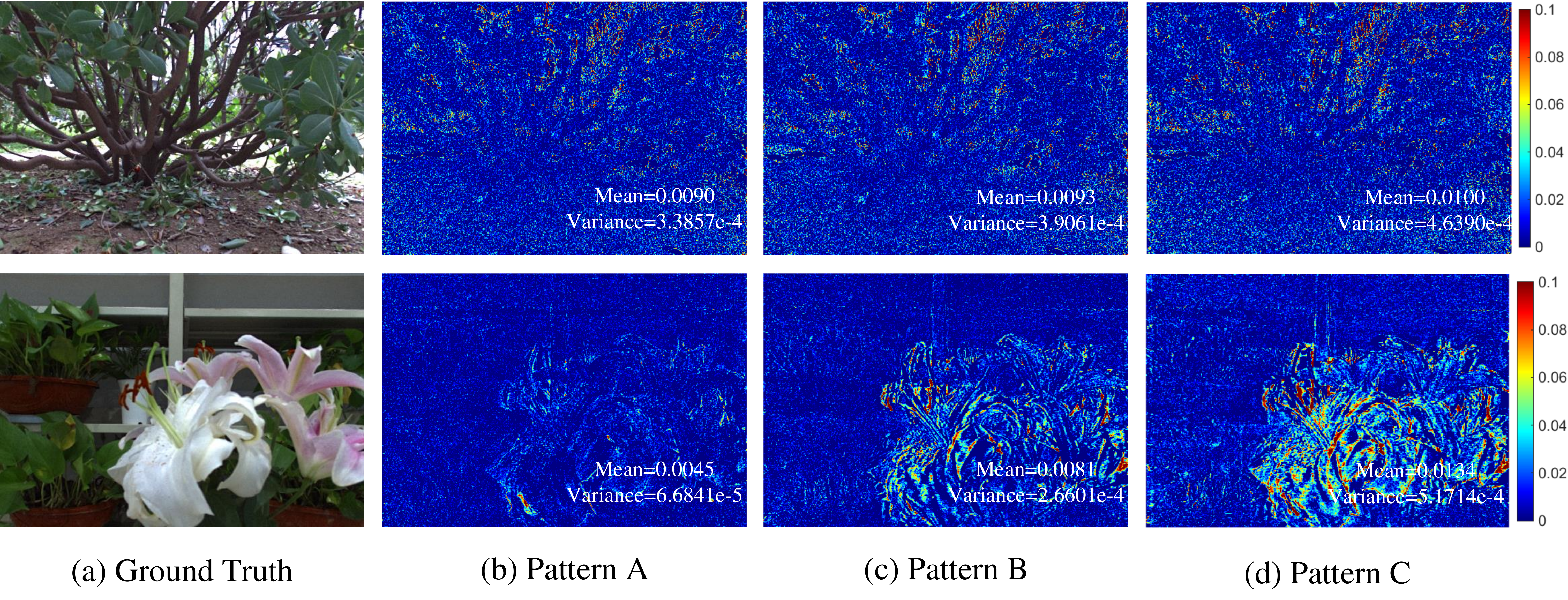}
\end{center}
\caption{ Reconstruction error analyses of different sampling patterns. (a) is ground truth view image. (b)-(d) are error maps and statistics under different light fields sampling patterns. }
\label{fig:samplerate}
\end{figure*}

Fig.~\ref{fig:views} shows view images in the reconstructed light field. The Toys scene contains plenty of textureless areas. Kalantari et al. \cite{kalantari2016learning}'s result shows heavy artifacts on the dog's mouth and the bottle, as shown in the blue and yellow boxes in the view image. The dog's mouth is teared up in their result while our result shows fidelity in these areas. The Basket scene is a challenging case due to the hollowed-out grids on the baskets. Plenty of occlusions are generated by the gridlines. The result of Kalantari et al. \cite{kalantari2016learning}'s method shows visual incoherency on grid area of baskets as shown in Fig.~\ref{fig:views}(b). The grids of the basket reconstructed by Wu et al. \cite{wu2017light}'s method are also twisted. Besides, the synthesized views by Kalantari et al. \cite{kalantari2016learning}'s method and Wu et al. \cite{wu2017light}'s method both show burring artifacts around the handle of basket, as shown in Fig.~\ref{fig:views}(b)(c). However, our results show higher permformance in those areas mentioned above. Moreover, our method has primely reconstructed high frequency details of the scenes. The Flower scene contains many leaves and petals with complex shapes which generates numerous occlusions. The results of Kalantari et al. \cite{kalantari2016learning} and Wu et al. \cite{wu2017light} show ghost effects around the petals and occlusion edges.  However, our method shows high accuracy in those occlusion and textureless areas, such as the place where two pedals with the same color overlap (see the yellow boxes of Flower scene in Fig.~\ref{fig:views}).

Fig.~\ref{fig:EPIs} shows the details of reconstructed EPI on Flower case and Basket case. For our method, the EPIs in Fig.~\ref{fig:EPIs} is cropped from our results. The EPI in Flower scene contains many thin tube formed by pixels from the flower's stamen. These thin tubes are mixed together in Kalantari et al. \cite{kalantari2016learning}'s result. The result of Wu et al. \cite{wu2017light}'s method shows cracked artifacts on EPI tubes. However, the EPI tubes in our results remain straight and clear. The Basket scene is a challenging case for EPI-based method. The grids on the basket also generate grids in EPI.  Therefore, EPI-based method can be challenged by the complex structures in EPI. According to the results, Wu et al. \cite{wu2017light}'s method shows many curved tubes in their result EPI. The result of Kalantari et al. \cite{kalantari2016learning} loses lots of details around tubes in EPI, while our result shows a structured EPI.

\begin{table}[!t]
\begin{center}
\centering
\caption{Comparison of reconstructing different number of novel light fields between two input light fields by our method. PSNR and SSIM values are averaged on all reconstructed view images of 6 testing scenes.}
\label{tab:tab2}
\begin{tabularx}{.7\textwidth}{C|C|C|C}
        \hline
                & Pattern A    & Pattern B       &Pattern C   \\
        \hline
        PSNR    & 39.10 & 37.08   &36.04 \\
        SSIM    & 0.952 & 0.938   &0.921 \\
     	\hline
\end{tabularx}
\end{center}
\end{table}

\subsubsection{Method Capacity.}
As shown in Fig.~\ref{fig:samplepattern}, with different sampling pattern, the number of light fields which need to be reconstructed between each pair of input light field is different. To test our method's capacity of reconstructing dense light field, we separately trained the network with different sampling patterns in Fig.~\ref{fig:samplepattern} to reconstruct 2, 3, 4 novel light fields between each two input light fields. Then, we evaluate the results over 6 testing scenes. Table.2 indicates that PSNR and SSIM values average on 6 testing scenes decrease as the reconstructing number increases between each pair of input light fields. Fig.~\ref{fig:samplerate} depicts $L_1$ error maps of the view images in reconstructed light field with different sampling patterns. It indicates that when the reconstructing number increases, the quality of reconstructed light fields also decreases. More results are shown in our supplementary material.



\section{Conclusions and Future Work}

We propose a novel learning-based method for reconstructing a dense light field from sparse sampling. We model novel light fields reconstruction as extending disparity consistency between dense and sparse light fields. Besides, we introduce a dense sampled light field dataset in which light field has the highest angular resolution so far. The experimental results show that our method can not only reconstruct a dense light field using 2 input light fields, but also extend to multiple input light fields. In addition, our method has a higher performance in the quality of synthesized view than state-of-the-art view synthesizing methods. 

Currently, our method is only able to deal with light fields which are captured along a sliding track without orientation change of principal axis. In the future, we will generalize our method to multiple degrees of freedom motion of light field camera. Furthermore, it would be interesting to choose suitable sampling rate of light rays automatically and reconstruct dense light field for a specific scene.

\bibliographystyle{splncs}
\bibliography{arxiv-submission}
\end{document}